\documentclass[conference]{IEEEtran}
\IEEEoverridecommandlockouts
\usepackage{cite}
\usepackage{amsmath,amssymb,amsfonts}
\usepackage{algorithmic}
\usepackage{hyperref}
\usepackage{graphicx}
\usepackage{footnote}
\makesavenoteenv{table}
\usepackage{textcomp}
\usepackage{caption}
\usepackage{subcaption}
\usepackage{balance}
\usepackage{xcolor}
\usepackage[T1]{fontenc}
\def\BibTeX{{\rm B\kern-.05em{\sc i\kern-.025em b}\kern-.08em
    T\kern-.1667em\lower.7ex\hbox{E}\kern-.125emX}}

\title{A Review of Nine Physics Engines for Reinforcement Learning Research}

\author{\IEEEauthorblockN{Michael Kaup\IEEEauthorrefmark{1},
Cornelius Wolff\IEEEauthorrefmark{1}, Hyerim Hwang,
Julius Mayer\IEEEauthorrefmark{7}, Elia Bruni\IEEEauthorrefmark{7}
\IEEEauthorblockA{Institute of Cognitive Science,
Osnabrück University\\
Osnabrück\\
Email: mkaup.research@gmail.com, cowolff@uos.de, hhwang@uos.de, research@jmayer.ai, elia.bruni@uos.de\\
\IEEEauthorrefmark{1} \textit{\footnotesize Shared first authorship} \\
\IEEEauthorrefmark{7} \textit{\footnotesize Shared senior authorship}}}}
\date{}

\begin{document}
\maketitle

\begin{abstract}
We present a review of popular simulation engines and frameworks used in reinforcement learning (RL) research, aiming to guide researchers in selecting tools for creating simulated physical environments for RL and training setups. It evaluates nine frameworks (Brax, Chrono, Gazebo, MuJoCo, ODE, PhysX, PyBullet, Webots, and Unity) based on their popularity, feature range, quality, usability, and RL capabilities. We highlight the challenges in selecting and utilizing physics engines for RL research, including the need for detailed comparisons and an understanding of each framework's capabilities. Key findings indicate MuJoCo as the leading framework due to its performance and flexibility, despite usability challenges. Unity is noted for its ease of use but lacks scalability and simulation fidelity. The study calls for further development to improve simulation engines' usability and performance and stresses the importance of transparency and reproducibility in RL research. This review contributes to the RL community by offering insights into the selection process for simulation engines, facilitating informed decision-making.
\end{abstract}

\begin{IEEEkeywords}
Reinforcement Learning, Physics, Engine, Review
\end{IEEEkeywords}

\section{Introduction}

Some of the more well-known research examples in reinforcement learning (RL) like Hide and Seek or the Sumo environment by OpenAI \cite{baker2020emergent, bansal2017emergent} involved embodied agents in simulated 3D environments \cite{Clay2021, adaptiveagentteam2023humantimescale}. According to Legg and Hutter \cite{legg2007intelligence} an agent's intelligence can be defined by its ``ability to achieve goals in a wide range of environments''. Taking this definition into account, the enablement of embodied agent-environment interaction is crucial. While deep learning (DL) libraries like TensorFlow \cite{tensorflow2015-whitepaper} or PyTorch \cite{paszke2019pytorch} and environment frameworks such as OpenAI Gym \cite{brockman2016openai} or PettingZoo \cite{terry2021pettingzoo} have lowered the entry barriers for RL research \cite{Zhou2020TowardsAD}, RL research is still held back by the difficulties of choosing and handling physics engines for implementing these interactions. In theory, multiple frameworks and simulation engines exist for RL experiments in physically simulated environments, yet RL researchers often do not describe the used simulation pipeline and their decision in detail (e.g. \cite{baker2020emergent,adaptiveagentteam2023humantimescale, park2023generative}). Thoroughly testing and comparing various simulation frameworks before choosing the right one is prohibitively time-intensive. Hence, researchers turn towards ready-made solutions but may struggle to find suitable resources and tools for creating RL environments, especially if they lack in-depth domain knowledge. Quantitative performance comparisons in the context of RL are rare in the existing literature and often provided by the engine developers themselves \cite{todorov2012mujoco, makoviychuk2021isaac}. Accordingly, these evaluations can be biased when it comes to the advantages and disadvantages of the presented engine. Of the available comparisons, many are outdated \cite{ivaldi2014tools, TassaYuval2015}, that is, they do not describe the current state and variety of frameworks accurately. Thus, current trends and developments in the field (e.g. increased need for environment complexity, multi-agent reinforcement learning, GPU-based simulation) are not sufficiently reflected in the literature. The remaining recent comparisons are largely focused on simulation capabilities for industrial robotics applications with RL \cite{bettini2022vmas, chen2022multirobolearn, körber2021comparing, collins2021review} or on the RL algorithms and specific environments \cite{ferigo2020ignition, kim2021survey}. A general and systematic review of the underlying engines, particularly one that also considers capabilities for multi-agent reinforcement learning (MARL) research, is missing.

\begin{figure*}
     \centering
     \begin{subfigure}[b]{0.49\textwidth}
         \centering
         \includegraphics[height=130pt]{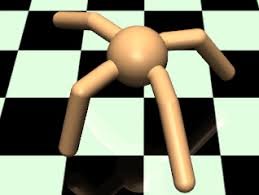}
         \caption{The ant environment in MuJoCo \cite{pateria2021end}}
         \label{fig:y equals x}
     \end{subfigure}
     \hfill
     \begin{subfigure}[b]{0.49\textwidth}
         \centering
         \includegraphics[height=130pt]{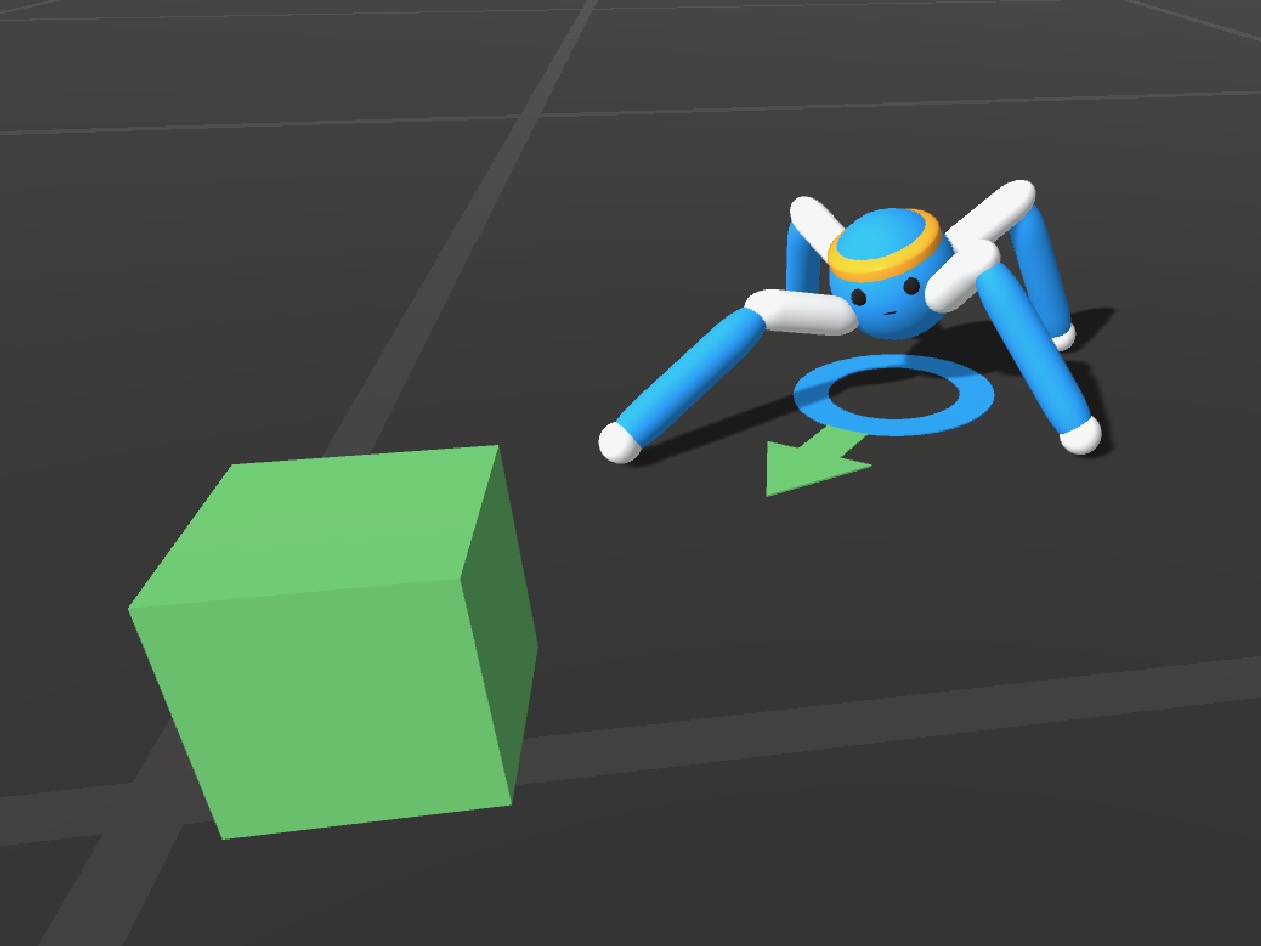}
         \caption{The crawler environment in Unity \cite{LearningUnity}}
         \label{fig:three sin x}
     \end{subfigure}
     \caption{Two similar environments realized in different engines}
\end{figure*}


Our review will focus on the usability of different engines for basic RL simulations to support researchers in choosing the appropriate creative tools for designing better and increasingly more challenging RL environments, algorithms, and training setups. \cite{juliani2020unity}. Therefore, we aim to review the most popular frameworks in RL research, regarding their popularity, feature range, feature quality, and usability. While most framework documentations offer simple environments like the one presented in Figure 1, we assess the engines' capabilities to generate multi-agent-ready 3D environments that can give rise to high task-complexity and task combinations. For this, we chose frameworks specialized for physics simulation (Brax, Chrono, Gazebo, MuJoCo, ODE, PhysX, PyBullet, and Webots) as well as the more broadly applicable game engine Unity. We provide insights into our selection process and briefly mention excluded candidates like Unreal Engine and Project Malmo in the \textit{Honourable Mentions} section.

The main contributions of this paper are:
\begin{itemize}
    \item A review of the engines' popularity in terms of citations.
    \item The evaluation of the engines' feature range, feature quality, and usability.
    \item A detailed assessment of the engines' RL and MARL capabilities.
\end{itemize}

\section{Methodology}

Our methodology comprises two major parts. First, we evaluate the popularity of different engines by looking at the number of citations of the respective publications, as well as the increase in that number of time. After that, we compare the physics engines along various relevant dimensions. 

\subsection{Popularity Analysis}

The popularity analysis aims to evaluate the popularity of physics engines for RL by performing an analysis of the citations of certain frameworks in scientific databases. We compared the popularity of the individual physics engines in the field of reinforcement learning research via the overall number of citations and the number of ML-related citations. We describe the specific steps carried out for the popularity analysis in the appendix.

\subsection{Feature Analysis}

The feature analysis aims to evaluate the feature range, quality, and usability of the physics engines for RL. We assess these criteria based on publications using these engines, the documentation provided by the engine's developers as well as previous papers reviewing performance and usability. We describe the specific steps carried out for the feature analysis in the appendix.

\subsection{Comparison Criteria}

We selected the following comparison criteria to reflect each framework's feature range, feature quality, and usability for RL research.

\begin{enumerate}
    \item Open source: Whether an engine is open-source, open-access, or closed-access. Open-source frameworks allow for greater accessibility, customization and better integration with external tools and existing frameworks. We did not consider any paid features.
    \item Documentation: The accessibility, extensiveness and visualization of the documentation, as well as the number and quality of provided examples.
    \item Community resources: The extent of relevant forum entries, Q\&As, and user-created models and environments.
    \item 3D Model library: The availability and capabilities of general-purpose RL agent models such as ants and humanoids.
    \item 3D Model creation: The ease of creating and customizing models for RL agents. Optimally, this is possible by manipulating objects within a well-interfaced editor. Model creation only via the handling of code in XML files and similar formats is rated unfavorably.  
    \item Environment library:  The availability and usefulness of general purpose 3D environments that can be used as training arenas for general purpose and RL.
    \item Environment creation: The ease of creating and customizing environments for RL agents. Optimally, this is possible by manipulating objects within a well-interfaced editor. Environment creation only via the handling of code in XML files and similar formats is unfavorable.   
    \item Sensors: The range of available sensors, e.g. camera, touch sensors, or radar.
    \item Gym Wrapper: The availability, usability, and feature range of gym wrappers for the particular engines. 
    \item Rigid body dynamics: The possibility and fidelity of simulating basic kinematic interactions. 
    \item Multi-joint dynamics: The possibility and fidelity of simulating complex multi-unit kinematic interactions in embodied agents.
    \item File formats: Support for importing and exporting Unified Robot Description Format (URDF) and MuJoCo Modeling XML File (MJCF). Both are XML file formats used for representing robot models and virtual agents. 
    \item Visualization: The graphical fidelity of the presentation of simulation episodes and results, availability of in-built rendering solutions, as well as the functional aesthetics and usability of the visualization interface.
    \item Performance: The optimization (or possibility of optimization) for training of RL agents by allowing for efficient parallel computing. This is a particularly important dimension, and we discuss the results in a dedicated \textit{Performance} section.
 \end{enumerate}

\section{Results}
\subsection{Popularity Analysis}

Table 1 shows the number of overall publications and ML-related publications citing each physics engine on the 6th of September 2023.

MuJoCo is the most popular physics engine in terms of citations, with over 3800 citations since its release in 2012. Gazebo follows closely, with 2698 citations since its release in 2004. Webots and PyBullet have also been well-cited, with over 988 and 1308 citations respectively. Brax, the newest of the engines listed, has received 166 citations since its release in 2021, which is relatively low but expected given its recent release. Unity's proportion of ML-related citations to overall citations might be skewed since the associated publication by \cite{juliani2020unity} addresses Unity as a platform for learning agents and is not an all-purpose introduction to the Unity engine as a game development toolkit. In this broader sense, Unity is more well-known.

\begin{table}[t]
\centering
\caption{Overall publications and ML publications citing each framework's original paper since it was first released}
\begin{tabular}{ |p{2cm}|p{1.5cm}|p{2cm}|p{1cm}| }
 \hline
 \multicolumn{4}{|c|}{Popularity Comparison}  \\
 \hline
 Physics Engine & Publications & ML Publications & since\\
 \hline
 Brax & 166 & 151 & 2021\\
 \hline
 Chrono & 170 & 104 & 2016\\
 \hline
 Gazebo & 2698 & 1948 & 2004\\
 \hline
 MuJoCo & 3827 & 3541 & 2012\\
 \hline
 ODE\footnotemark[1] & 1573 & 143 & 2004\\
 \hline
 PhysX & 310 & 288 & 2021\\
 \hline
 PyBullet\footnotemark[1] & 1308 & 1000+ & 2016\\
 \hline
 Unity & 576 & 528 & 2018\\
 \hline
 Webots & 988 & 548 & 2004\\
 \hline
\end{tabular} \\
\end{table}

\setcounter{footnote}{1}
\footnotetext[1]{Counted using Google Scholar}

\begin{figure}[b]
\includegraphics[width=.5\textwidth]{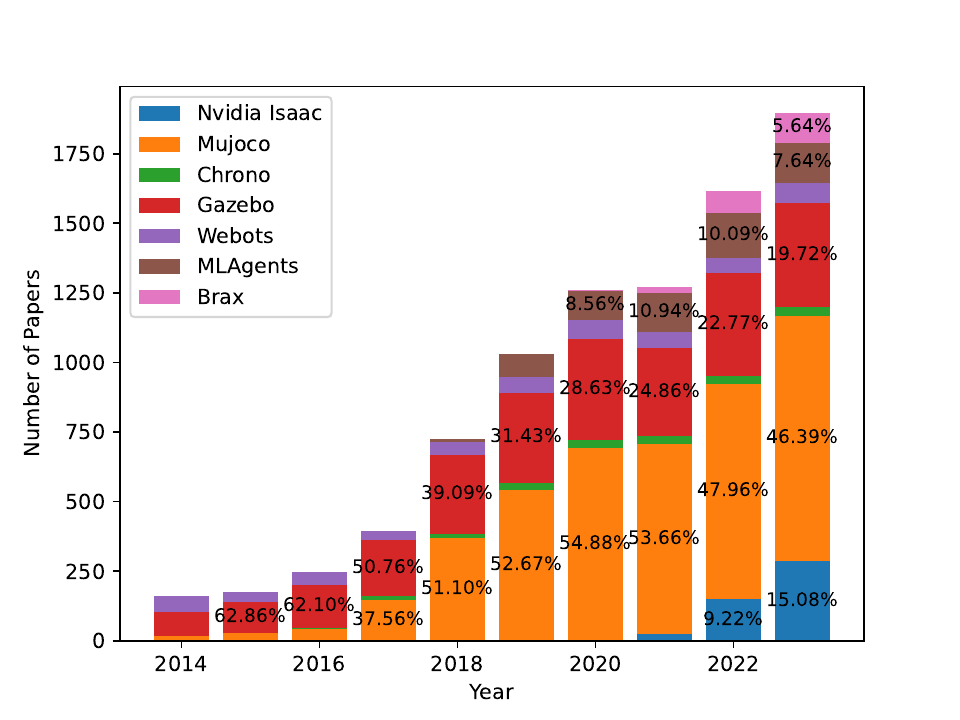}
\caption{Yearly citations of the frameworks' original publications}
\end{figure}

In terms of the number of ML papers that have used a particular physics engine, MuJoCo, and PyBullet are the most popular, with over 3541 and 1000+ papers citing them, respectively. Unity and Gazebo have also been widely used in ML, with over 528 and 1948 citations each. Brax, ODE, PhysX, and Webots have been used in a smaller number of ML papers, ranging between 143 and 548 citations. 

Figure 2 shows, that the usage of Nvidia Isaac has seen substantial growth from 2021 to 2023, with mentions increasing from 20 to 286. Mujoco maintained its popularity from 2020 to 2023, with mentions remaining high at 695 and 880, respectively. Chrono's usage has remained relatively small in comparison but stable over the years, with only minor fluctuations in mentions from 2016 to 2023. Gazebo experienced fluctuations in usage, but overall retained a stable number of citations between 374 in 2023 and 324 in 2019. Webots had a consistent number of mentions from 2011 to 2023, with 72 mentions in 2023. ML Agents saw a significant increase in mentions from 2019 to 2022, with a peak of 163 mentions. However in 2023, there was a slight decrease to 145 mentions. Brax's usage has steadily been increasing over recent years to 107 mentions 2023 since its publication in 2021. These year-to-year differences in usage reflect the evolving preferences and trends within the robotics and RL research communities.

It is worth noting that the popularity of a physics engine can be influenced by factors such as ease of use, documentation, community support, and compatibility with other tools. Therefore, while the number of citations and ML papers is a useful measure of popularity, it may not necessarily reflect the best physics engine for a particular application.

\subsection{Feature Analysis} 
\begin{table*}[t]
\begin{center}
\caption{Feature and Usability Comparison (full description can be found in Chapter II).  Legend: feature or a range of features is fully available and functional (+ +), feature is available, but lacking in some regards (+), feature is either available, but lacking or only available via workarounds ($-$), feature is not available or difficult to integrate ($-$ $-$)}
\begin{tabular}{ |p{3cm}||p{1,2cm}|p{1,2cm}|p{1,2cm}|p{1,2cm}|p{1,2cm}|p{1,2cm}|p{1,2cm}|p{1,2cm}|p{1,2cm}| }
 \hline
 \multicolumn{10}{|c|}{Feature and Usability Comparison} \\
 \hline
 Features & Brax & Chrono & Gazebo & MuJoCo & ODE & PhysX & PyBullet & Unity & Webots\\
 \hline
 Open Source & + + & + + & + + & + + & + + & + & + + & $-$ & + +  \\
 Documentation & $-$ $-$ & $-$ & $-$ & + & $-$ & + + & $-$ & + + & + +\\
 Community resources & $-$ $-$ & $-$ & + & + & $-$ & $-$ & $-$ & + + & + \\
 Model library & + & + + & + + & + + & $-$ $-$ & + & + + & + + & +\\
 Model creation & $-$ & $-$ & $-$ & $-$ & $-$ $-$ & $-$ & $-$ & + + & +\\
 Environment library & $-$ $-$ & $-$ $-$ & + & $-$ $-$ & $-$ $-$ & $-$ & $-$ $-$ & + + & +\\
 Environment creation & $-$ $-$ & $-$ $-$ & $-$ $-$ & $-$ $-$ & $-$ $-$ & $-$ $-$ & $-$ $-$ & + + & $-$\\
 Visualization & + + & + + & + + & + & $-$ $-$ & + + & + & + + & + + \\
 Gym Wrapper & + & $-$ & + & $-$ & $-$ $-$ & + + & + & + + & + +  \\
 MARL capabilites & $-$ & $-$ & $-$ & + + & $-$ & + & + + & + & $-$ \\
 Rigid body dynamics & + + & + + & + + & + + & + +  & + + & + + & + + & + + \\
 Multi-joint dynamics & + + & + + & + & + + & + + & + + & + + & $-$ & + \\
 Sensors & $-$ & + + & + + & + + & $-$ $-$ & + + & $-$ & + +& + +  \\
 URDF support & + + & $-$ & + + & + + & $-$ $-$ & + + & + + & $-$ $-$ & + \\
 MJCF support & + + & $-$ & + + & + + & $-$ $-$ & + + & + + & + & $-$  \\
 \hline

\end{tabular}
\end{center}
\end{table*}

\subsubsection{MuJoCo}
\textit{MuJoCo} (Multi-Joint dynamics with Contact; \cite{todorov2012mujoco}) is an open-source physics simulation engine specialized for robotics, biomechanics, animation, and ML. It is owned and curated by Google DeepMind and has become popular with leading RL researchers, such as OpenAI's multi-agent research \cite{openai2019learning, baker2020emergent}. It provides rigid body dynamics in interaction with their environment, such as collision detection and contact resolution, support for various joint types as well as actuation options. This makes MuJoCo especially suitable for RL focussed on embodied movement. The simulation can be visualized interactively with a native graphical user interface (GUI) for rendering the simulation including meshes and textures, but not while training. Advanced rendering options like complex lighting and shaders are limited, but less relevant for RL. Furthermore, it allows users to selectively run parts of the computation pipeline for flexibility \cite{todorov2012mujoco}. No direct gym environment integration is provided. MuJoCo can be accessed via DeepMind's Control Suit (dm\_control; \cite{tassa2018dmcontrol}). However, this Python API is currently poorly documented and lacks transparency as well as usability. Accessing objects via dm\_control can be difficult. We encountered bugs and unexpected behaviors. Furthermore, it restricts the creation and customization of XML models. MuJoCo has decent documentation that is only hard to navigate because of its large extent and lacking structure. The documentation itself is well presented with highlighted code, images, videos, and GIFs. Python bindings are taught in Google Colab notebooks. An overview and demo notebook is provided, but many functionalities found on the GitHub repository are not explained. Along with the engine, Google DeepMind offers a collection of pre-defined, importable models equipped with joints and limbs that are useful for embodied RL\footnote{\url{https://github.com/deepmind/mujoco\_menagerie}}. Models and environment content can be defined and customized in XML files. The resulting physical model can be hard to pre-visualize from just code. Hence, additional 3D modeling tools that can export MJCFs or URDFs, like \textit{Blender}, may be required for custom model creation in more complex projects. MuJoCo natively runs on a single thread, but multi-threading can also be implemented\footnote{\url{https://mujoco.readthedocs.io/en/latest/programming/simulation.html}}. Furthermore, there are multiple libraries like Envpool \cite{weng2022envpool}, which enable users to increase the sampling performance significantly by applying highly optimized vectorization techniques. At the same time it has to be mentioned that such libraries are often only optimized for classic single-agent gym tasks and are not directly usable for MARL setups. If such libraries can be used for custom multi-agent environments, they often require proficiency in programming languages like C or a generally deep understanding of the used framework. However, single thread performance is sufficient for the wide range of tasks and often not worth the additional core usage \cite{todorov2012mujoco}. Despite its daunting entry barrier and lacking documentation, the range of features make MuJoCo a powerful and flexible framework for RL.



\subsubsection{PyBullet}

\textit{PyBullet} \cite{pyBbllet2016} is an open-source Python module for robotics simulation and ML that allows users to dynamically create and simulate physics-based environments for RL. PyBullet wraps the C-API of Bullet and offers simple integration with TensorFlow and PyTorch. PyBullet supports loading URDFs and MJCFs with dedicated functions \cite{pyBbllet2016}, which can be used to import and implement ant, humanoid, half-cheetah, and similar models as shown in the PyBullet Quickstart Guide \cite{coumans2021pybullet}, \cite{PyBulletQuickstartGuide}. Notably, shapes and multi-body models cannot only be defined in external XML formats, but also directly via PyBullet functions. Users can equip agents with sensors to capture information such as position, orientation, velocity, or contact forces. Complex 3D environments are not provided. PyBullet has functional visualization but is not specialized for graphics rendering. Complex lighting, textures, and shaders are not supported \cite{juliani2020unity}. PyBullet does not provide a prebuilt MARL environment. However, \cite{panerati2021learning} developed an open-source OpenAI Gym-like environment called \textit{gym-pybullet-drones} for multiple quadcopters. Several researchers utilized this framework to conduct MARL research \cite{feng2022joint, schmidt2022introduction, trang2023multi}, thus providing evidence for the capabilities of the underlying PyBullet engine in principle. PyBullet's documentation is hard to access as the main site links to three different sources, which are limited to Google Docs and poorly formatted PDF files without code highlighting on their GitHub repository. This makes the needed information scattered and hard to connect. The main document, the PyBullet Quick-start Guide, provides somewhat extensive information over 75 pages that lacks in-depth use cases. Example applications and showcases are found on GitHub, however not in Python. Nevertheless, PyBullet has a large community \footnote{\url{https://pybullet.org/Bullet/phpBB3/}}.



\subsubsection{Unity}
\textit{Unity} \cite{juliani2020unity} is a popular game development engine. In contrast to the other presented engines, Unity is not open-source, but rather open-access where not all features are available in the free version. Paid plans for professional and entrepreneurial use exist. Unity stands out compared to the other engines, due to its intuitive interface that combines all features in a single workspace. With Unity ML Agents, it offers a large open-source toolbox with 3D training arenas, model assets that are already equipped with RL algorithms, like Proximal Policy Optimization (PPO) and Soft-Actor Critic (SAC), that work out-of-the-box. Through its asset store, Unity offers a large array of official and community-built packages, that can be especially useful for the design of various environments. Many of these are free and continuously updated. Unity ML Agents also offers a Python API to integrate externally defined agents. Unity physics, the engine's package for deterministic rigid body dynamics simulation, can be complemented with plug-ins for the engines \textit{Havok} \footnote{\url{https://docs.unity3d.com/Packages/com.havok.physics@0.1/manual/index.html}} and \textit{MuJoCo} \footnote{\url{https://mujoco.readthedocs.io/en/latest/unity.html}}. A wide range of sensors is available. Unity has highly accessible and extensive documentation, with well-structured tables of content and hyperlinks to related sections. Code examples are well-highlighted and embedded in visually appealing tutorials that cover all aspects of the engine.

Many RL paradigms implant agents into video-game-like scenarios, where they have to solve tasks similar to those set for human players \cite{juliani2020unity}. Historically, some of the most notable milestones of AI research have been performances in games. This includes digital versions of classical board games, like chess and go \cite{silver2017chess} as well as established video games, such as StarCraft II \cite{vinyals2019starcraft} and Dota 2 \cite{openai2019dota}. Furthermore, the emergence of generalizable skills in agents that are applicable to a range of different video games and RL environments is one of the core objectives of much of RL research \cite{legg2007intelligence, schrittwieser2020muzero}. This has been tried and tested successfully \cite{Clay2021} with AI benchmarks based on Unity, such as the Obstacle Tower Challenge \cite{juliani2019obstacle}. 
For these reasons, Unity should, in theory, be the natural choice of engine for the implementation of any video-game-like RL scenario. However, the fact that Unity is specialized for game development poses several disadvantages. Unity's optimization for video games clashes with the RL training demands of maximizing frames, i.e. simulation steps per unit of time and computational resource \cite{ward2020usingunity}. Another hurdle is that Unity ML Agents is only convenient as long as the whole pipeline is assembled within Unity. There are significant hurdles when it comes to integrating a Unity environment into existing Python code. The limited development possibilities on top of Unity as opposed to within Unity can be identified as a core problem. Unity makes the setup of multi-agent scenarios quite practical and easily implementable. However, efficiency becomes even more of a problem for MARL than for single-agent training. If a simulation becomes too complex and computationally expensive, Unity increases the time between simulation frames. This hurts simulation fidelity and constrains MARL approaches, as MARL has typically a high amount of interacting units, especially with embodied setups. Workarounds to manually fix simulation fidelity and training efficiency problems exist (see \cite{ward2020usingunity}). Despite these disadvantages, Unity is used by leading researchers for complex and computationally demanding RL scenarios \cite{openendedlearningteam2021, adaptiveagentteam2023humantimescale}. However, both did not utilize the ML Agents toolkit but went for custom solutions based on \cite{ward2020usingunity}. Google DeepMind's extensive resources have to be considered here, as this adaptation of Unity to specific RL needs might not be as easily imitated. 



\subsubsection{Gazebo}

Gazebo \cite{koenig2004gazebo} is an open-source robot simulation software for simulating and testing robotic systems developed by Open Robotics. It is the official simulation platform for the DARPA Robotics Challenge \cite{ivaldi2014tools}. Gazebo offers rigid body dynamics, various types of joints, and sensors through multiple supported physics engines, including ODE, Bullet, Simbody, and DART, allowing users to easily switch between them. Users can utilize a wide range of sensors. Gazebo provides a wide range of pre-built models and environments designed for simulation purposes. With its own editor system, users can create and modify simple models directly in the GUI. The Gazebo GUI renders the 3D simulation in real time. Gazebo allows users to define and customize robot models using URDF or SDF. However, customization of a large environment could take a lot of time \cite{ivaldi2014tools}. Gazebo provides the Python package sdformat-mjcf\footnote{\url{https://github.com/gazebosim/gz-mujoco/tree/main/sdformat_mjcf}} that allows bidirectional conversion between SDF and MJCF. 
Gazebo's documentation consists of a tutorial section with explanatory images, examples, highlighted code, and some hidden automatically generated documents. The rudimentary are not explained at all in the documents, only somewhat in the tutorials. No Python bindings are explained or available apart from PyGazebo\footnote{\url{https://github.com/jpieper/pygazebo}} \footnote{\url{https://pygazebo.readthedocs.io/en/latest/pygazebo.html}} and Ignition\footnote{\url{https://gazebosim.org/api/gazebo/2.10/index.html}}, where it is unclear to the user whether the information provided is official.

Gazebo does not provide an official gym wrapper. However, the Gazebo simulator offers a rich set of APIs and tools for simulation, physics-based modeling, and visualization, which can be used alongside the OpenAI Gym framework by creating a custom gym wrapper. There is open-source project called \textit{gym-gazebo2} \cite{lopez2019gymgazebo2} that provides a gym wrapper specifically designed for integrating Gazebo simulations with RL algorithms. Gym-Ignition \footnote{\url{https://ignitionrobotics.org}} is a framework that provides reproducible robotic environments for RL and robotics research \cite{ferigo2020ignition}.
 Users can create environments in either Python or C++. This feature combined with the multitude of supported engines enables effective randomization and helps prevent potential overfitting issues. Gym-Ignition currently has limited support for photorealistic rendering \cite{ferigo2020ignition}. Although Gazebo itself does not provide an environment for setting up MARL, \textit{MultiRoboLearn} \cite{chen2022multirobolearn} provides a framework to apply MARL to Gazebo, specialized for robotics simulation. Base Gazebo exhibits considerable performance loss with multi-agent setups \cite{bettini2022vmas}. Gazebo's problematic usability makes the implementation of 3D environments difficult. However, users can trade-off simulation speed and computational cost for higher fidelity. Thus, it seems more appropriate for robotics RL, especially industrial applications \cite{körber2021comparing}.

\subsubsection{PhysX/IsaacGym}

Nvidia's \textit{PhysX} \cite{physx2020} is an SDK mainly used for visual effects, video game development, robotics and medical simulation\footnote{\url{https://developer.nvidia.com/blog/introducing-isaac-gym-rl-for-robotics/}}. Using Nvidia IsaacGym \cite{makoviychuk2021isaac} as a gym environment, PhysX can also run RL algorithms in its virtual environment. Examples given by IsaacGym are implemented in PyTorch, but TensorFlow is equally feasible. While PhysX is open source, IsaacGym is not, which might hinder its customization \cite{ferigo2020ignition}. Typical MuJoCo and RL Games\footnote{\url{https://github.com/Denys88/rl_games}} models can be used. With IsaacSim in Nvidia's Omniverse, an even more specialized toolkit for robotics simulation exists. IsaacGym provides a PPO implementation and supports MJCF and URDF. PhysX supports photorealistic rendering in an intuitive interface. Range, contact, force and camera sensors are available via extensions\footnote{\url{https://docs.omniverse.nvidia.com/app_isaacsim/app_isaacsim/manual_isaac_extensions.html}}. PhysX and Isaac Gym are excellently documented with a digestible structure, visual examples, extensive documents, explanatory text as well as video tutorials and GIFs. 

IsaacGym's distinguishing feature is that it leverages GPU acceleration to increase simulation speed compared to other engines' CPU-based physics simulation. By directly connecting the simulation backend with PyTorch Tensors, IsaacGym aims to avoid CPU bottlenecks. If CPU power availability is an issue, this can be an immense advantage, as it potentially increases the number of RL environments that can run simultaneously on a single computer and decreases the need for costly computing clusters. Notably, ant, humanoid and hand movement benchmarks showed decreased training time \cite{makoviychuk2021isaac}. Nevertheless, GPU-based simulation can be hindering to successful RL research as the GPU will often have to be fully dedicated to running the deep learning algorithm and the CPU is rarely fully occupied in MARL. Thus, PhysX might be more of a specialized tool for robotics RL and less suitable to basic RL research. At the same time, even though there are some examples of MARL setups in Nvidia Isaac \cite{chen2022towards}, the implementation requires more in-depth programming knowledge than other comparable setups. However, as PhysX and IsaacGym represent one of the few high-usability, unified frameworks for RL and physics simulation \cite{ferigo2020ignition} at the moment, the drawback might in some scenarios be worth the cost.

\subsubsection{ODE}

\textit{ODE}\footnote{\url{https://www.ode.org/}} (Open Dynamics Engine) provides access to an open-source C/C++ library designed for simulating rigid body dynamics. It supports advanced joint types and integrated collision detection with friction. It is commonly used for simulating vehicles and dynamic objects in 3D environments. The documentation is scattered across several different web pages and is hard to navigate. The single-page user manual and a dedicated tutorial section provide explanations of core functionalities and automatically generated documents with a severely dated appearance are provided. Some rather short code examples without highlighted code are hidden within ODE's GitHub repository. The FAQ on GitHub is very thorough, however. Many features relevant to the criteria evaluation were not locatable or not documented.
 While ODE does not provide a Python API directly, there exists PyODE\footnote{\url{https://pypi.org/project/PyODE/}}, which is a set of open-source Python bindings for the Open Dynamics Engine. ODE does not provide direct support for URDF or MJCF format. Additionally, ODE does not include built-in sensor functionalities. Visualization of simulation results as well as the interface in which it is embedded were neither high-resolution nor up to modern UI standards. Overall, ODE is outdated and unwieldy on the usability side and makes for a strenuous implementation of state-of-the-art RL paradigms. Furthermore, it has little relevance in today's RL research literature (see popularity comparison). Therefore, ODE seems only applicable to current RL research setups through its comparatively more modern front-ends and engine integrations in Gazebo and Webots.




\subsubsection{Webots}

\textit{Webots} \cite{webots2004} is a widely used open-source robot simulation software developed by Cyberbotics, supporting C, C++ as well as Python. It simulates a wide range of robotic systems, relying on a customized version of the ODE 3D dynamics library. Webots makes highly specific sensors available, from camera and touch sensors to radar and lidar. Its GUI offers real-time 3D visualization and a front-end for modifying simulation models. Webots allows a robot controller to export URDFs. However, generated URDFs are currently limited to a few elements such as the definition of a box, cylinder, or sphere. Webots does not directly support MJCFs. It has its own native file format, PROTO, for defining the structure, appearance, and dynamics of robot models. The documentation of Webots is well-structured, providing user and installation guides that are easy to access. Its documentation makes good use of images, videos, and code chunks with highlighting. Both the reference manual and the user guide are quite extensive. They have a dedicated tutorials section that is extensive with great explanations, code, and images. Since Webots is built on top of ODE, users will have some inconvenience in checking the poorly structured ODE documentation for certain parameter or function explanations. \\
The Webots environment library is limited to a few specific examples, such as an apartment and a factory. The available Webots model library is specialized for complex robotics simulation\footnote{\url{https://www.cyberbotics.com/doc/guide/robots?version=R2019a-rev1}} rather than general purpose RL. Simple models for embodied RL, like the typical ant, are possible to implement in Webots, but have to be made from scratch or imported as a third-party asset. Similarly, base Webots offers no integration for Tensorflow or PyTorch as well as no multi-agent simulation capabilities, but \textit{Deepbots}\footnote{\url{https://github.com/aidudezzz/deepbots}} \cite{kirtas2020deepbots} closes these gaps. Deepbots interfaces Webots with OpenAI Gym and adds functionalities necessary for controlling RL agents and gym environments while hiding Webots features that are not relevant for RL. Thus, the RL algorithm backend, TensorFlow or PyTorch is connected with the simulation side. However, Deepbots, as the name suggests, is specialized for robotics, and the complexity of the provided environments is achieved through complicated multi-joint robotics models, rather than tightly packed 3D worlds. Several simple ready-to-use environments, such as CartPole, PitEscape, and FindBall \footnote{\url{https://github.com/aidudezzz/deepworlds}}, can be used to benchmark RL algorithms in Webots \cite{kirtas2020deepbots}. However, no MARL algorithmic environments are provided \cite{chen2022multirobolearn}. Deepbots has not caught on yet with the RL research community (see citations of \cite{kirtas2020deepbots}). Generally, Webots appears to not lend itself to highly scalable training and therefore MARL \cite{bettini2022vmas}, as it runs each simulation in its GUI and can only be parallelized by opening multiple instances of Webots manually \cite{körber2021comparing}. Its high-fidelity simulation and user-friendly GUI, however, make it especially suitable for robotics RL setups that do not have high parallelization demands.


\subsubsection{Brax}

\textit{Brax}, "a differentiable physics engine for large scale rigid body simulation" \cite{freeman2021brax} is an open-source physics simulation engine written in JAX that is accessible via Google Colab \footnote{\url{https://colab.research.google.com/github/google/brax/blob/main/notebooks/basics.ipynb}}. Brax simulates physical systems made up of rigid bodies, joints, and actuators and offers high flexibility for creating multi-agent environments with different physics properties, observation spaces, and action spaces. \cite{freeman2021brax}. It is specifically designed for RL and optimized to efficiently run parallel physics simulations alongside the RL algorithm on a single accelerator. Brax specifically aims to solve similar problems and offer similar models to MuJoCo. Whereas most aforementioned simulation frameworks separate simulation (CPU) and RL algorithm (GPU/TPU), Brax brings both together on a single GPU or TPU chip in order to reduce latency. Brax is quite new and poorly documented. The documentation comprises only a short readme file and three example notebooks in Google Colab. No central webpage for information is available. No models or example environments are provided. Furthermore, community resources, like assets or helpful forum entries, are not to be found. Brax's model library offers implementations of the basic MuJoCo models, such as the ant, humanoid, and half-cheetah\footnote{\url{https://ai.googleblog.com/2021/07/speeding-up-reinforcement-learning-with.html}} \footnote{\url{https://github.com/google/brax}}, but not much beyond. No complex training environments are provided. According to \cite{bettini2022vmas}, Brax has problems with complex MARL, precisely because of its computationally expensive high-fidelity simulation. Scaling the number of agents increases this problem and after a threshold of only a low number of agents the simulation reaches a standstill. Its main selling point, GPU-based simulation, is also offered by PhysX/IsaacGym with a better feature range and usability. For these reasons, Brax in its current form does neither seem to be a platform for general RL, nor fill a more specific niche. Despite these criticisms, we recognize this innovative approach and the effort to make deep learning more accessible and less reliant on high-performance clusters.

\subsubsection{Chrono}

\textit{Chrono}\footnote{\url{https://projectchrono.org/}} \cite{Chrono2016} is an open-source modeling and physics simulation engine for robotics and vehicle dynamics. It offers a wide range of physical simulation capabilities, including collision detection, rigid body dynamics, and various force elements. PyChrono\footnote{\url{https://api.projectchrono.org/development/pychrono_introduction.html}} \cite{PyChrono2022} wraps the C++ simulation library and allows users to build physical models and exchange data between the simulation and ML framework. For RL setups, Chrono provides a custom PyTorch PPO implementation \cite{PyChrono2022}. A Chrono-based simulation environment to design and test end-to-end exists \cite{EndToEndChrono2022}. However, it is mostly focused on training autonomous vehicles and robots in off-road settings \cite{PyChrono2022, young2022enabling}. Gym Chrono\footnote{\url{https://github.com/projectchrono/gym-chrono}} is a set of PyChrono-based OpenAI Gym environments. Gym Chrono provides examples for training via TensorFlow and PyTorch. \textit{Chrono::Sensor}\footnote{\url{https://api.projectchrono.org/manual_sensor.html}} provides a rich set of sensor modules which can simulate cameras, lidars, radars, gyroscopes etc. It does not directly support URDF or MJCF format natively and its model library mainly offers vehicles and robots. However, in Gym Chrono, users can utilize ant models \cite{benatti2020training} for RL setups. Chrono provides a limited environment library. Its GUI provides convenient control and monitoring of simulations. Also, Chrono integrates with various visualization libraries, such as Irrlicht, OpenGL, and Unity3D, to render the simulated systems in run-time. Chrono's and PyChrono's documentation is comprised of a poorly structured automatically generated document. The main document is somewhat extensive, but lacks explanation of fundemantals, while the dedicated tutorial section is code-only and does not explain anything on a conceptual level. Only sparse images and no explanatory videos are provided. We found Chrono's negligible relevance in the ML literature (see popularity comparison), poor usabilty and focus on vehicle robotics \cite{collins2021review} to indicate a limited usefulness as an engine for RL research and MARL purposes.

\subsubsection{Honorable mentions}

\textit{Unreal Engine} is a popular open-access game development engine. Recently, Unreal Engine introduced \textit{Learning Agents}, a plugin geared towards game developers who want to write AI bots. The Learning Agents API can be accessed via Unreal Engine's general user interface and can be used with C++ and Python. Agents can be trained with an existing PPO algorithm. Support for SAC and Q-Learning is provided. However, the Learning Agents API has been available for less than seven months as of December 2023 and has correspondingly not been widely cited in the relevant RL literature. For this reason and because Epic Games, the developers of Unreal Engine, state themselves, that Learning Agents is not a general purpose ML framework, we won't go into detail comparing it to other engines. Third-party tools for RL with Unreal Engine, e.g. \textit{Mindmaker}, are available via the Unreal Engine Marketplace.
\textit{Godot} is a open source game development engine that can be used for RL research via the framework \textit{Godot RL Agents} \cite{beeching2021godot}.
 However, Godot itself is not widely used and has even less relevance for RL \cite{beeching2021godot}. More interesting for RL researchers is Generally Intelligent's \textit{Avalon} \cite{albrecht2022avalon}, a 3D simulator based on Godot that lets users plug RL agents into ready-made environments with complex task interaction possibilities. 
\textit{Project Malmo} \cite{johnson2016malmo} is a useful platform for exploration-related RL experimentation that is based on the \textit{Minecraft} engine. As such, it is constrained by the limitations of the underlying video game \cite{juliani2020unity} and cannot provide complex, embodied physics simulation with high fidelity and it cannot be used to build scenarios that are not feasible in \textit{Minecraft}. Similar limitations are true for \textit{ViZDoom} \cite{kempka2016vizdoom} which is based on the underlying engine of the popular video game \textit{Doom} and \textit{DeepMind Lab} \cite{beattie2016deepmindlab}, based on \textit{Quake III}. \textit{VMAS} (Vectorized Multi-Agent Simulator; \cite{bettini2022vmas}) is a 2D physics engine written in PyTorch that is specifically designed with efficient MARL in mind. However, the lack of 3D implementations severely limit the possible complexity of the training environment as well as the agent-environment interaction. 

\section{Performance}

\cite{mohammed2020reinforcement} showed that MuJoCo is better than PyBullet and ODE at generalizing learning to other engines, i.e. agents who learned to solve a task in MuJoCo still perform when the same task is transferred and implemented in a different engine. Agents trained via PyBullet did not transfer their learning at all. Thus, it might be the case that, for example, agents trained on PyBullet just learn to navigate PyBullet environments, whereas agents trained on MuJoCo learn to navigate any similarly simulated environment. MuJoCo's developers \cite{TassaYuval2015} compared the speed, simulation stability, and simulation accuracy of Bullet, MuJoCo, ODE, and PhysX by implementing the same scenario in each engine and measuring the time steps at which simulation errors occurred. They found MuJoCo to have the best performance out of all engines, especially in scenarios that simulate bodies with many joints or connected elements. \cite{körber2021comparing} implemented a similar broad range of use cases with Gazebo, MuJoCo, PyBullet, and Webots and compared the ratio of simulation time that can be achieved in real-world time (RTF). MuJoCo was reported to have a high RTF across scenarios, at the cost of some accuracy. PyBullet achieved a lower RTF but was highlighted for its superior usability. Meanwhile, Gazebo was found to be unwieldy and most suitable for simulations that are intended to be transferred to real systems. Webots showed high stability and RTF even in the most complex scenarios but is criticized for its lack of native parallelization support. As already established, Brax scales poorly in MARL setups \cite{bettini2022vmas}.

\section{Limitations}
To rigorously assess and compare the quantitative performance of the presented frameworks, one would have to implement the same scenarios for typical RL use cases in all engines. This goes beyond the scope of this paper and due to the sheer required effort has not been attempted to a sufficient degree by any other publication to the best of our knowledge. For statements on technical details of the engines we relied on information from the engine publishers and developers, as well as external researchers who used and evaluated them. Therefore, the performance evaluation is neither exhaustive nor compares all frameworks on equal footing. Correspondingly, the evaluation might be skewed by the availability of data on the engines. On the other hand, sparse information is a legitimate shortcoming.



\section{Conclusion}

In this paper, we looked at 9 frameworks for RL research and reviewed them regarding their popularity, feature range, feature quality and usability and we contributed to the field by providing an overview of the engines that enables researchers to make informed decision when choosing their framework for RL simulation. We paid special attention to the engine's MARL capabilities. We conclude, that for successful RL research, it is first necessary to sharply define the intended scenario and research whether a suitable implementation is not already available. For example, there is no reason to handle the usability inconveniences of MuJoCo if it is sufficient to have a MARL setup in 2D. This holds especially true for RL training on video game scenarios, where the selection of benchmarks is plentiful. For anything more specific, the choice of physics engines naturally depends on the defined needs and available resources of the project. 

MuJoCo is currently the dominant framework for RL research due to its good performance and flexibility, even though its documentation is sometimes lacking and might make usage for smaller teams more difficult than with other competitors. Compared to the other engines, MuJoCo currently provides one of the best foundations for MARL due to its high simulation fidelity and high training efficiency. Nevertheless, the creation of complex training environments for MuJoCo can be comparatively strenuous. Notably, high-fidelity simulation might not be useful for all training setups, as it can massively increase the computational demands while adding little benefit to setups where accurate kinematics is not paramount. PyBullet offers similar features and usability as MuJoCo, but consistently rates worse in performance reviews \cite{mohammed2020reinforcement, TassaYuval2015, körber2021comparing}. For this, it makes up in a wide range of dedicated functions for loading and defining objects and models. Once the user has disentangled the documentation, RL scenarios are straightforward to implement in PyBullet.


While designing an environment is the easiest in Unity out of all frameworks, Unity is not optimized for parallel computing and large-scale training. Unity has various pre-implemented MARL scenarios and can support simple multi-agent interactions, but has problems with scaling complexity and simulation fidelity. One should also consider that low simulation fidelity impacts the reproducibility of results negatively \cite{ward2020usingunity}. Unity's and Unity ML Agents' strong suit is the RL implementations of video-game scenarios. Beyond this purpose, Unity seems most suitable for proofs of concept or RL experiments that are not intended to scale the training beyond a certain threshold. Right now, Brax fails to impress, due to its limited available resources and documentation and poor multi-agent performance. However, Brax is quite new and might be updated with more useful features in the near future. PhysX/IsaacGym, on the other hand, excels in terms of usability and provides a unified framework for scenario creation, simulation, and RL. Both Brax and IsaacGym rely on GPU-driven simulation which can be disadvantageous for large-scale RL research. Base ODE is outdated both in terms of feature range and usability and accordingly has limited impact on current RL research, while Chrono lacks important features such as URDF and MJCF support. We found, that Gazebo and Webots represent powerful tools for high-fidelity simulation robotics with decent usability. However, both are not geared towards MARL applications.

Custom creation of complex environments and corresponding libraries with pre-built solutions remain a gap in available simulation pipelines. Another research gap is the lack of technical training performance comparison for MARL in complex 3D environments as well as the implementation difficulty for typical scenarios in each engine. Further development and research is needed in these areas. Study-specific transparency and reproducibility remain a structural problem in RL research, with many leading institutes and research teams opting for closed access. Further guidance on environment creation and replication and better usability of the relevant tools is thus strongly necessary. Symptomatic for the field, the most performant engine (MuJoCo) has poor usability and the most user-friendly engine (Unity) suffers from poor performance. For significant progress in the field, a better combination of the best of the two worlds has to be achieved.

\subsection{Author contributions}
This research was completed within the scope of the MicrocosmAI project\footnote{\url{https://microcosm.ai/}}, which made this project possible. M.K. made the main writing contribution and organized the research and writing process. C.W. contributed to the methodology, crawling algorithm and popularity comparison. H.H. made contributions to the Chrono, Gazebo, ODE and Webots chapters. J.M., E.B. and the larger MicrocosmAI research project contributed expertise, feedback and a framework for supervision. All authors researched data and literature and contributed substantially to the conceptualization of the submitted version.

\section{Acknowledgements}
This work was funded by the Deutsche Forschungsgemeinschaft (DFG, German Research Foundation) — 456666331.

\balance
\begin{scriptsize}
\bibliographystyle{IEEEtranS}
\bibliography{references.bib}
\end{scriptsize}

\section{Appendix}
\subsection{Popularity Analysis}

\begin{enumerate}
\item Database selection: To carry out the popularity analysis, we chose the \textit{Semantic Scholar} database, as it is commonly used in the field of RL and offers a free-to-use API to download the meta-data of all citations \cite{Kinney2023TheSS}. 
\item Gathering lists of citations: From the databases, we downloaded the meta-data of all papers, which cited the original papers introducing the physics engines.
\item Selection criteria: We defined selection criteria to filter the crawled citations. We limited our search to papers published between 2016 and 2022 with a focus on research in the domain of RL or ML in general. For this, we filtered the results by the keywords ``Reinforcement Learning'', ``Machine Learning'', ``Artificial Intelligence'' or ``Training'' in either the paper title, abstract, or keywords. For papers that were not available on \textit{Semantic Scholar}, we manually counted all of their ML-related citations on \textit{Google Scholar}. This limits comparability to some degree, which is why manually counted papers were not included in Figure 2 where the yearly changes in citations are displayed.
\item Data analysis: We performed a quantitative analysis of the extracted data and compared the results in terms of the overall number of citations and the number of ML-related citations between the individual physics engines.
\end{enumerate}

\subsection{Feature Analysis}

\begin{enumerate}
\item Review selection: We searched for existing reviews and papers related to the topic of our study. We used search queries relating to RL, embodied RL, MARL, and the names of the individual physics engines. Specifically, we used the following search term for each of the physics engines: "\textless Physics engine name\textgreater  AND ('Machine Learning' OR 'ML' OR 'Reinforcement Learning' OR 'RL' OR 'Artificial Intelligence' OR 'AI' OR 'Multi-Agent')"
\item Selection criteria: The relevance of the retrieved papers to this review was determined by the number of citations, the topic, the publication date as well as the used methods and perceived quality of the research.
\item Data extraction: We reviewed the data from the selected papers, including the RL algorithm used, the number of agents, the evaluation metrics, and the results.
\item Data analysis: We performed an analysis of the extracted data, including thematic review. Based on these results, we conducted a comparative analysis to identify the strengths and weaknesses of each physics engine.
\end{enumerate}

\end{document}